\documentclass[conference]{IEEEtran}
\IEEEoverridecommandlockouts

\usepackage{filecontents}
\usepackage{cite}
\usepackage{amsmath,amssymb,mathtools,amsfonts}
\usepackage{algorithmic}
\usepackage{graphicx}
\usepackage{textcomp}
\usepackage{xcolor}
\usepackage{hyperref}
\usepackage[tableposition=top,labelsep=period,font=footnotesize]{caption}

\def\BibTeX{{\rm B\kern-.05em{\sc i\kern-.025em b}\kern-.08em
    T\kern-.1667em\lower.7ex\hbox{E}\kern-.125emX}}
\begin{document}

\bstctlcite{IEEEexample:BSTcontrol}

\title{Total Recall: Understanding Traffic Signs using Deep Hierarchical Convolutional Neural Networks\\
\thanks{Code is available at \href{https://github.com/Sourajit2110/DilatedSkipTotalRecall}{https://github.com/Sourajit2110/DilatedSkipTotalRecall}}
}

\author{\IEEEauthorblockN{Sourajit Saha\IEEEauthorrefmark{1}, Sharif Amit Kamran\IEEEauthorrefmark{2} and Ali Shihab Sabbir\IEEEauthorrefmark{3}}
\IEEEauthorblockA{\textit{Center for Cognitive Skill Enhancement} \\
\textit{Independent University Bangladesh}\\
Dhaka, Bangladesh \\
sourajit@iub.edu.bd\IEEEauthorrefmark{1}, sharifamit@iub.edu.bd\IEEEauthorrefmark{2}, asabbir@iub.edu.bd\IEEEauthorrefmark{3}}
}

\maketitle

\begin{abstract}
Recognizing Traffic Signs using intelligent systems can drastically reduce the number of accidents happening worldwide. With the arrival of Self-driving cars it has become a staple challenge to solve the automatic recognition of Traffic and Hand-held signs in the major streets. Various machine learning techniques like Random Forest, SVM as well as deep learning models has been proposed for classifying traffic signs. Though they reach state-of-the-art performance on a particular data-set, but fall short of tackling multiple Traffic Sign Recognition benchmarks. In this paper, we propose a novel and one-for-all architecture that aces multiple benchmarks with better overall score than the state-of-the-art architectures. Our model is made of residual convolutional blocks with hierarchical dilated skip connections joined in steps. With this we score 99.33\% Accuracy in German sign recognition benchmark and 99.17\% Accuracy in Belgian traffic sign classification benchmark. Moreover, we propose a newly devised dilated residual learning representation technique which is very low in both memory and computational complexity.
\end{abstract}


\section{Introduction}
As Self-driving cars are being introduced in major cities, intelligent traffic signs recognition has become an essential part of any autonomous driver-less vehicles \cite{robustdetection,videotrack,vehiclelocal,driveracitivity}. Transitioning from a vehicle with driver to a driver-less vehicle should come in steps. 
The major issue seems to be the person driving the vehicle not being attentive enough to notice the traffic signs in due time. A safety harness should be placed which automates the recognition of traffic signs and alerts the driver in due time as they might suffer from fatigue or other causes \cite{precrash,fatigue}. Solving this can greatly reduce the number of casualty in major roads and highways. Orthodox computer vision techniques \cite{sift} and machine learning based architectures  were
popular for traffic sign classification \cite{svm,mogolvision,realsigncolor,plsa}, yet failed miserably to state-of-the-art deep learning architectures in recent times. Currently, deep convolutional neural networks are capable of outperforming any traditional machine learning methods for traffic sign recognition. Though many deep learning models have been proposed for traffic sign classification \cite{capsule-traffic,rogue-sign-adversial}, they failed to provide sufficient evidence regarding overall performance for multiple traffic sign benchmarks. However, some of them scored nearly Top-1\% accuracy for a particular benchmark like GTSRB data-set \cite{gtsrb-1,gtsrb-2}, yet no further example of their models were shown for other popular benchmarks like BelgiumTS data-set \cite{btsrb-1,btsrb-2,btsrb-3} or LisaTSR benchmark \cite{lisa}. Hence, it is evident that there is no single architecture which outperforms multiple benchmarks for Traffic Sign Recognition. It is important to mention that, a model should not be embedded on an autonomous vehicle before proving its accuracy and performance for multiple benchmark. In this paper, we propose a novel deep convolutional neural network architecture. It is accompanied by hierarchical structure with customized skip connections \cite{fcn,skip} in steps. The skip connections are made with dilated convolutions \cite{dilation} with different filter sizes to increase the receptive field of the feature extractors.
With this architecture we reach 99.33\% accuracy in German Traffic Sign Recognition data-set \cite{gtsrb-1,gtsrb-2}. On the other hand, we reach 99.17\% accuracy in Belgian Traffic Sign data-set \cite{btsrb-1,btsrb-2,btsrb-3}. Concurrently, it can be embedded to any autonomous vehicle as its reaches competitive scores on multiple benchmark.

\begin{table}[bp]
\centering
\caption{Architecture Comparison with Large-Scale Classifiers on GTSRB.}
\begin{tabular}{|c|c|c|c|c|}
\hline
Architecture & \begin{tabular}[c]{@{}c@{}}Top-1\\ Error, \( \epsilon_{1} \)\end{tabular} & \begin{tabular}[c]{@{}c@{}}Top-5\\ Error, \( \epsilon_{5} \)\end{tabular} & \begin{tabular}[c]{@{}c@{}}Parameters\end{tabular} & \begin{tabular}[c]{@{}c@{}}Training\\ Time\end{tabular} \\ \hline
\textbf{Our} & \textbf{0.67\%} & \textbf{0.34\%} & \textbf{6.256 M} & \textbf{2.03 Hr} \\ \hline
Pre-Resnet-1001 \cite{he2016identity} & 0.71\% & 0.47\% & 10.2 M & 6.07 Hr \\ \hline
Resnet-50 \cite{he2016deep} & 0.73\% & 0.52\% & 23.8 M & 4.55 Hr \\ \hline
\end{tabular}
\label{tab1}
\end{table}

\section{Literature Review}
Various methodical approaches were adopted for recognizing traffic signs in the field of computer vision. However, most of them were cling to only one set of benchmark. In this work, we discuss such methodologies starting from classical brute force approaches to modern learning representations. Most of them tackle the fundamental classification, detection and localization challenges surrounding traffic sign recognition. 

\subsection{Feature Extractor as Classification Technique}
Classical approaches include techniques such as Histogram Oriented Gradients (HOG) \cite{hog,hog-color}, Scale Invariant Feature Transformation (Sift) \cite{sift} and Sliding Window \cite{sliding-window}. HOG based techniques were used for visual salience \cite{hog} and then for color exploitation for pedestrian detection\cite{hog-color}. Moreover, gradients of RGB images were computed along with different normalized, weighted histograms for finding the best detection algorithm to find pedestrians and signs. Furthermore, scale Invariant Feature Transform (SIFT) \cite{sift-traffic} technique was used for classification of traffic signs, whereas sliding window \cite{sliding-window} approach was used to find candidate ROIs (Region of interest) within a small-sized window, and then further verified within a large-sized window for higher accuracy in object detection. 

\begin{figure}[tp]
\centering
\includegraphics[width=6.5cm,height=5.5cm]{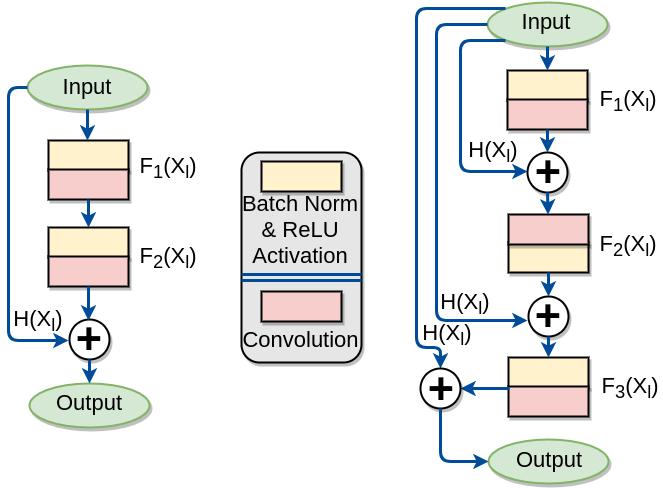}
\caption{Residual skip connections. Left: A conventional pre-activation residual skip block with one residual function. Right: [Inner Residual Skip Block] Our proposed pre-activation residual skip block with two residual functions.}
\label{fig1}
\end{figure}

\subsection{Orthodox Machine Learning Techniques}
Machine learning methods like support vector machines \cite{svm,svm-2}, linear discriminant analysis (LDA) \cite{LDA}, kd-trees \cite{kd-tree}, maximally stable extremal regions\cite{mser} and Random forest \cite{random-forest} swept away the brute force approaches in traffic sign recognition. Concurrently, LDA is based on  maximum likelihood estimate or maximum posteriori estimation between classes and class
densities are represented by multivariate Gaussian and common co-variance matrix \cite{LDA}. However, discriminant function analysis is very similar to logistic regression, and both can be used to answer the same research questions \cite{spss}. Logistic regression does not have as many presumptions and restrictions as linear discriminant analysis. However, when discriminant analysis’	supposition are met, it is more superior and stronger than logistic regression \cite{stat}. In Random Forest a set of non pruned random decision trees are used to make an ensemble architecture through which the best classification scores are achieved \cite{random-forest}. The decision trees are made with features selected randomly from the training set. For traffic sign recognition test data is validated by all the decision trees and the categorical output and probability scores are based on the majority voting. Support Vector Machines(SVM) is used for classification problem and it classifies the data by dividing the n-dimensional data plane with a hyper plane\cite{svm-2}. SVM can transform the classification plane to higher dimensions using kernel trick \cite{kerneltrick}. The method separates the non-linearly scattered data using a non-linear kernel function. The major problems for above techniques are, features needs to be hand-engineered and machine learning is heavily dependent on human-intervention \cite{mogolvision,random-forest,LDA}. These approaches can not handle variable length images neither can converge better with data augmentation and low pre-processing.

\subsection{Deep Learning Approaches}
To get rid of the drawback of the above mentioned techniques, new architectures based on
deep learning algorithms were emerged \cite{pcnn,stdnn,hlsgd,mcdnn}.The reasons being the increase in the amount of available computing resources and access to huge annotated data. Currently, almost all the state-of-art architectures for traffic signs are Convolutional neural networks. The first of its kind was the introduction of LeNet architecture\cite{lenet} for traffic sign recognition for German Traffic Sign data-set Benchmark Challenge \cite{gtsrb-1,gtsrb-2}.
What makes convolutional neural networks more accurate and easily implementable is it's tower like structure that can process information and learn features in depth. There are variations in each block and layers, like convolution layer is the main feature extractor which uses filters with small receptive field to process input\cite{imagenet}, pooling layer is used for reducing spatial dimension \cite{pooling} and then there is dense or fully-connected layer which takes input from all the neurons of the previous layers and shares the information to connected layers \cite{neuralnetwork}. A loss function is defined which is reduced by Back-propagation \cite{backpropagation}. Moreover, a new type of convolution called dilated convolution \cite{dilation} has been replacing vanilla convolution in latest architectures. The main intuition behind this is it increases receptive field exponentially if stacked on top of each other \cite{parsenet}, whereas in vanilla convolutional layer, stacking increases receptive field linearly \cite{fcn,neuralnetwork}.

\begin{figure}[tp]
\centering
\includegraphics[width=6cm,height=5.4cm]{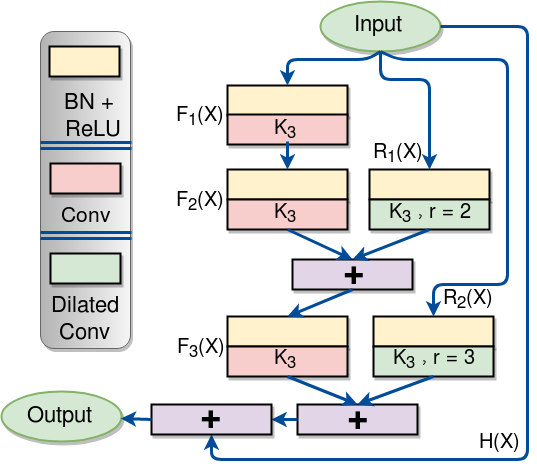}
\caption{Proposed Dilated Inner Residual Skip Block.}
\label{fig2}
\end{figure}

\section{Proposed Methodology}

\subsection{{Skip Connections in Convolutional Neural Networks}}
With the introduction of deep learning, the computer vision research community have been thriving to acquire precision level accuracy in image classification along with other computer vision tasks. Concurrently, we have seen deeper architectures like \cite{simonyan2014very} with 143.6 Million parameters and improvements like \cite{szegedy2016rethinking} that goes even deeper with parameters as many as 23.8 Million. However with the depth increasing, the accuracy does not idealistically get better. As the study \cite{srivastava2015highway}, \cite{he2015convolutional} suggests, accumulating more kernels causes performance degradation without provoking the high variance - high bias paradigm. Clearly, the recursive process of learning more features based on the previously learned features is not always optimal. Therefore, there has to be a trade-off between how much to learn from the already learned features with respect to the immediate previous layer and layers further beyond. To answer that intriguing question \cite{he2016deep} first introduced deep residual learning, where each building block learns new features and simultaneously passes the previous layer as they are. This skip like connections gives the network the feasibility to attain features from layers further beyond when needed, affording the network to learn more interesting features with juxtaposition of features from any previous layers at any point in time, also known as the identity mapping. Further analysis of identity mapping \cite{he2016identity} has confirmed that using batch normalization and activation before a convolutinal layer in each residual block increases the classification performance. As shown in Figure ~\ref{fig1}, a conventional residual block \cite{he2016identity} has 2 pre-activation convolutions in it's residual branch, the accumulation of which we denote by \( F^{R}(X_{l}) \) and identity mapping, \( H(X_{l}) = X_{l} \). The conventional Residual operations are demonstrated in Equation \eqref{1}. In stead of using a bunch of these residual blocks we tried using a fewer with newly designed skip connections that are capable of learning more interesting features. Inside each residual block we embedded two more residual units, which goes inside the two composite functions as illustrated in Figure ~\ref{fig1}, therefore acting as one single residual block with two more residual blocks within. We call it Inner Residual Block and we denote it's output function as \( F^{IR}(X_{l}) \). The computation of Inner Residual Block is illustrated by Equation \eqref{2}, \eqref{3}, \eqref{4}.
\begin{equation}
X_{l+1} = F^{R}(X_{l}) + H(X_{l}) = F_{2}( F_{1}( X_{l} ) ) + H(X_{l}) \label{1}
\end{equation}
\begin{equation}
 F^{IR}(X_{l}) = F_{3}( F_{2}( F_{1}( X_{l} ) + H(X_{l}) ) + H(X_{l}) ) \label{2}
\end{equation}
\begin{equation}
X_{l+1} = F^{IR}(X_{l}) + H(X_{l}) \label{3}
\end{equation}
\begin{equation}
X^{IR}_{L} = X_{l} + \sum^{L-l}_{k=l} F_{3}( F_{2}( F_{1}( X_{k} ) + H(X_{k}) ) + H(X_{k}) )  \label{4}
\end{equation}

\begin{figure}[tp]
\centering
\includegraphics[width=8cm,height=3.7cm]{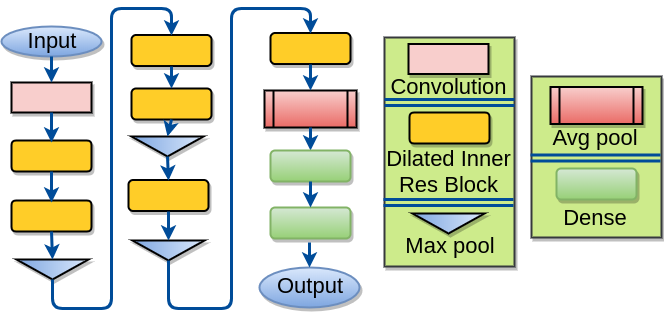}
\caption{Proposed Architecture with Dilated Inner Residual skip Connections.}
\label{fig3}
\end{figure}

\subsection{{Dilated Convolutional Operation}}
Dilated convolutional operation has been a recent development that has lead to improved classification \cite{dilation}, \cite{parsenet} and segmentation \cite{amit2017efficient} performances. A dilated convolutional kernel uses the same number of operation as that of a regular kernel yet it skips the tensor by a fixed number of pixels, thus resulting in a extended receptive field. A regular convolution filter of size \( ( k \times k ) \), where \( k = \alpha \), \( \forall (\alpha > 1 \cap \alpha \in \mathbb{Z}) \); has a receptive filed of \( E_{1} = ( k \times k ) \) whereas, a dilated convolution kernel with size of \( ( k \times k ) \) and dilation rate \( r = \beta \), \( \forall (\beta > 1 \cap \beta \in \mathbb{Z}) \); has a receptive field of \( E_{2} = ( (k + (k-1)(r-1) ) \times ( (k + (k-1)(r-1)) ) \). Equation \eqref{5} elaborates the extension in receptive field with no added computation where, \( \Delta E > 0 \). We denote the number of parameters with a regular and dilated kernel for the \( l^{th} \) layer with depth, \( D^{[l]} \) as \( P_{1} \) and \( P_{2} \) respectively. A comparative analysis of parameter requirement is formulated using Equation \eqref{6}, \eqref{7}, \eqref{8}.
\begin{equation}
\Delta E = (k-1)(r-1)(2k + (k-1)(r-1)) \label{5}
\end{equation}
\begin{equation}
P_{1} = (k+(k-1)(r-1))^{2} \times D^{[l]} \times D^{[l-1]} \label{6}
\end{equation}
\begin{equation}
P_{2} = (k)^{2} \times D^{[l]} \times D^{[l-1]} \label{7}
\end{equation}
\begin{equation}
P_{1}/P_{2} = \Big[ 1 + \frac{(k-1)(r-1)}{k} \Big]^{2} \label{8}
\end{equation}
Since \( (k-1)(r-1) \geq k \) therefore \( P_{1} > P_{2} \), denoting that dilated convolution confirms a reduction in parameters. Given a input tensor of spatial dimension \( (X \times X) \), a deep convolutional neural network takes 1 regular kernel with zero padding and 1 stride to reduce the spatial dimension to \( ( (X-(k-1)) \times (X-(k-1)) ) \), whereas, the same kernel with dilation rate r can reduce the tensor dimension to \( ( (X-(k-1) \times r ) \times (X-(k-1) \times r ) ) \). Intrinsically, running the same regular kernel r times would result in such spatial reduction.

\begin{table}[tp]
\centering
\caption{Inspection of the Dilated Inner Residual CNN Architecture.}
\begin{tabular}{|c|c|c|}
\hline
Layer name & \begin{tabular}[c]{@{}c@{}}Output Size\end{tabular} & \begin{tabular}[c]{@{}c@{}}Kernel Operation\end{tabular} \\ \hline
Conv1 & \( 56 \times 56 \times 64 \) & \( 1 \times 1 \), 64, stride = 1 \\ \hline
Conv2a,b & \( 56 \times 56 \times 64 \) & \( D_{64} \times 2 \) \\ \hline
Pool1 & \( 28 \times 28 \times 64 \) & \( 3 \times 3 \) Max Pool, stride = 2 \\ \hline
Conv2c & \( 28 \times 28 \times 64 \) & \( D_{64} \times 1 \) \\ \hline
Conv3a & \( 28 \times 28 \times 128 \) & \( D_{128} \times 1 \) \\ \hline
Pool2 & \( 14 \times 14 \times 128 \) & \( 3 \times 3 \) Max Pool, stride = 2 \\ \hline
Conv4a & \( 14 \times 14 \times 256 \) & \( D_{256} \times 1 \) \\ \hline
Pool3 & \( 7 \times 7 \times 256 \) & \( 3 \times 3 \) Max Pool, stride = 2 \\ \hline
Conv4b & \( 7 \times 7 \times 256 \) & \( D_{256} \times 1 \) \\ \hline
AvgPool & \( 1 \times 1 \times 256 \) & \( 7 \times 7 \) Average Pool \\ \hline
Dense1 & 512 & Fully Connected \\ \hline
Dense2 & Number of Classes & Fully Connected + Softmax \\ \hline
\end{tabular}
\label{tab2}
\end{table}

\subsection{{Inner Residual Dilated Skip Convolution}}
Figure ~\ref{fig2} depicts how we collocate both the dilation and residual unit concepts together. To posit the inner mechanism, we denote \( k_{m}= m \), where m represents both height and width of kernel k. In each of our block we apply two \(k_{3}\) kernels to the input tensor in the first place. Then we skip a step and apply a \(k_{3}\) kernel with dilation, r= 2 and get to the same state that had used 2 kernels, which then gets added and passed into another \(k_{3}\) kernel. Again we apply a \(k_{3}\) with dilation rate, r= 3 thus scraping the layout of using 3 kernels, which then we add with the \(3^{rd}\) \(k_{3}\) kernel results; ready to go as a output. However, we add a identity mapping before outputting the tensor and denote the dilated convolution function as \(R(X_{l})\), residual output as \( F^{DIR}(X_{l}) \), number of parameters for non dilated inner residual unit and dilated inner residual unit as \( P_{l}^{IR} \) and \( P_{l}^{DIR} \) respectively. Equation \eqref{9}, \eqref{10}, \eqref{11}, \eqref{12} elaborates the detailed calculations of Dilated Inner Residual Blocks.
\begin{equation}
 F^{DIR}(X_{l}) = F_{3}( F_{2}( F_{1}( X_{l} ) + R_{1}(X_{l}) ) + R_{2}(X_{l}) ) \label{9}
\end{equation}
\begin{equation}
X_{l+1} = F^{DIR}(X_{l}) + H(X_{l}) \label{10}
\end{equation}
\begin{equation}
X^{IR}_{L} = X_{l} + \sum^{L-l}_{k=l} F^{DIR}(X_{k})   \label{11}
\end{equation}
\begin{equation}
P_{l}^{IR}-P_{l}^{DIR} = D^{[l]} \times D^{[l-1]} \times ( ( K_{5}^{2} - K_{3}^{2} ) + ( K_{7}^{2} - K_{3}^{2} ) ) \label{12}
\end{equation}
Since \( k_{7}>k_{5}>k_{3} \) therefore \( P_{l}^{IR}>P_{l}^{DIR} \), ergo our architecture assures a stringent reduction in parameters. Studies \cite{saha2018lightning}, \cite{cai2017fractal} have shown that divide and merge feature mapping helps learn coarse features, thus we have incorporated the idea in each of the dilated inner residual blocks. Figure ~\ref{fig3} construes our network architecture while Table \ref{tab2} interprets all the kernel operations and filter output dimensions. As each blocks are capable of simultaneously looking at the same tensor with different view points and retain all of the cor-relational tensors together therefore we address this approach as \textbf{``Total Recall"}. Furthermore, all the regular and dilated convolutions in all dilated inner residual blocks use the same number of kernels. The term \( D_{\Omega} \) used in Table \ref{tab2} refers to the fact that all the kernels in a particular block uses \( \Omega \) of channels.

\begin{figure}[tp]
\centering
\includegraphics[width=8cm,height=2cm]{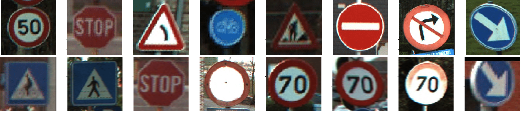}
\caption{Different traffic signs from Belgium Traffic Sign Classification Data-set}
\label{fig4}
\end{figure}

\begin{table}[bp]
\centering
\caption{Different Data-set}
\begin{tabular}{|c|c|c|c|c|c|}
\hline
Dataset & \begin{tabular}[c]{@{}c@{}}Total\\ images\end{tabular} & \begin{tabular}[c]{@{}c@{}}Training\\ image\end{tabular} & \begin{tabular}[c]{@{}c@{}}Test\\ image\end{tabular} & \begin{tabular}[c]{@{}c@{}}Image\\ Size\end{tabular}       & \begin{tabular}[c]{@{}c@{}}Number of\\ Classes\end{tabular} \\ \hline
GTSRB   & 51839                                                  & 39209                                                    & 12,630                                               & \begin{tabular}[c]{@{}c@{}}15x15 to\\ 250x250\end{tabular} & 43                                                          \\ \hline
BTSC   & 7095                                                   & 4575                                                     & 2520                                                 & \begin{tabular}[c]{@{}c@{}}11x10 to\\ 562x438\end{tabular} & 62                                                          \\ \hline
\end{tabular}
\label{tab3}
\end{table}

\section{Experimentation}

\subsection{Data-set Preparation}
We train and test our models on two sets of benchmark. The first one is German Traffic Sign Benchmark or GTSRB \cite{gtsrb-1,gtsrb-2} and the second is Belgium Traffic Sign Benchmark, also known as BTSC \cite{btsrb-1,btsrb-2,btsrb-3}. In this way, we get a viable model which performs very well on arbitrary data. There is essentially no difference between the data-sets except for the number of classes. All the RGB images from both the data-sets are cropped to $56\times56$ spatial dimension for both test and training. Table \ref{tab3} details all the information related to GTSRB and BTSC data-sets. Figure ~\ref{fig4} illustrates a preview of the BTSC data-set.

\begin{table}[tp]
\centering
\caption{Architecture Comparison for GTSRB}
\begin{tabular}{|c|c|}
\hline
Models                 & Top-1 Accuracy (\%) \\ \hline
\textbf{Ours}          & \textbf{99.33} \\ \hline
PCNN \cite{pcnn}                  & 99.3          \\ \hline
$\mu Net$ \cite{munet} & 98.9          \\ \hline
Human \cite{gtsrb-2}                 & 98.84         \\ \hline
HOG  \cite{btsrb-1}        & 96.14         \\ \hline
CDNN \cite{mcdnn}                   & 98.5          \\ \hline
Multi-scale-CNN        & 98.31         \\ \hline
pLSA  \cite{plsa}                 & 98.14         \\ \hline
Capsule NN  \cite{capsule-traffic}           & 97.62         \\ \hline
Random Forest  \cite{random-forest}        & 96.14         \\ \hline
\end{tabular}
\label{tab4}
\end{table}

\begin{table}[bp]
\centering
\caption{Architecture Comparison for BTSC}
\begin{tabular}{|c|c|}
\hline
Models        & Top-1 Accuracy (\%)  \\ \hline
\textbf{Ours} & \textbf{99.17} \\ \hline
HOG \cite{btsrb-1}          & 98.34          \\ \hline
OneCNN \cite{onecnn}       & 98.17          \\ \hline
\end{tabular}
\label{tab5}
\end{table}

\subsection{The Neural Net Training and Performance Evaluation}
\begin{equation}
    \mathcal{L}(\hat{y},y) = - \frac{1}{N} \sum^{N}_{j} [ y_{j} \log(\hat{y}_{j}) + (1-y_{j}) \log(1-\hat{y}_{j}) ] \label{13}
\end{equation}
\begin{equation}
    \alpha^{[i]} =
    \begin{cases}
        \alpha^{[i-1]} & \text{if \( \mathcal{L}^{[i-1]} \leq \mathcal{L}^{[i-2]} \leq \mathcal{L}^{[i-3]} \) } \\
        0.1 \alpha^{[i-1]} & \text{if \( 0.1 \alpha^{[i-1]} \leq 10^{-12} \)} \\
        \alpha^{[i-1]} & \text{otherwise}
    \end{cases}
    \label{14}
\end{equation}
While training, we split the 39,209 GTSRB images into 1226 batches and the 4,575 BTSC images into 143 batches, each with 32 images per batch. We have used categorical cross-entropy loss function, \( \mathcal{L} \) to calculate approximation error using Equation \eqref{13}, where \( \hat{y} \), y denotes prediction, true label correspondingly. Concurrently, to optimize the approximation we used Adam \cite{kingma2014adam} with a initial learning rate \( \alpha = \) \( 10^{-4} \) that updates itself using Equation \eqref{14} with reference to validation loss at \( i^{th} \) iteration. We trained both data-sets separately for 30 epochs on a 8 Gigabyte Nvidia GeForce GTX 1070 GPU using keras \cite{chollet2015keras}, a deep learning library. The GTSRB data-set was trained for 2.03 hours while the BTSC data-set took 0.23 hour to train. We ended up with the best performance (Table \ref{tab4}) on GTSRB benchmark as well as (Table \ref{tab5}) on BTSC benchmark. Figure ~\ref{fig6} shows the validation accuracy and loss plots after every iterations which depicts how quick the validation loss decreases while keeping a consistent accuracy, thus demonstrates the robustness of the proposed architecture (6.256 Million Parameters). 

\begin{figure*}[tp]
\centering
\includegraphics[width=17.8cm,height=6.5cm]{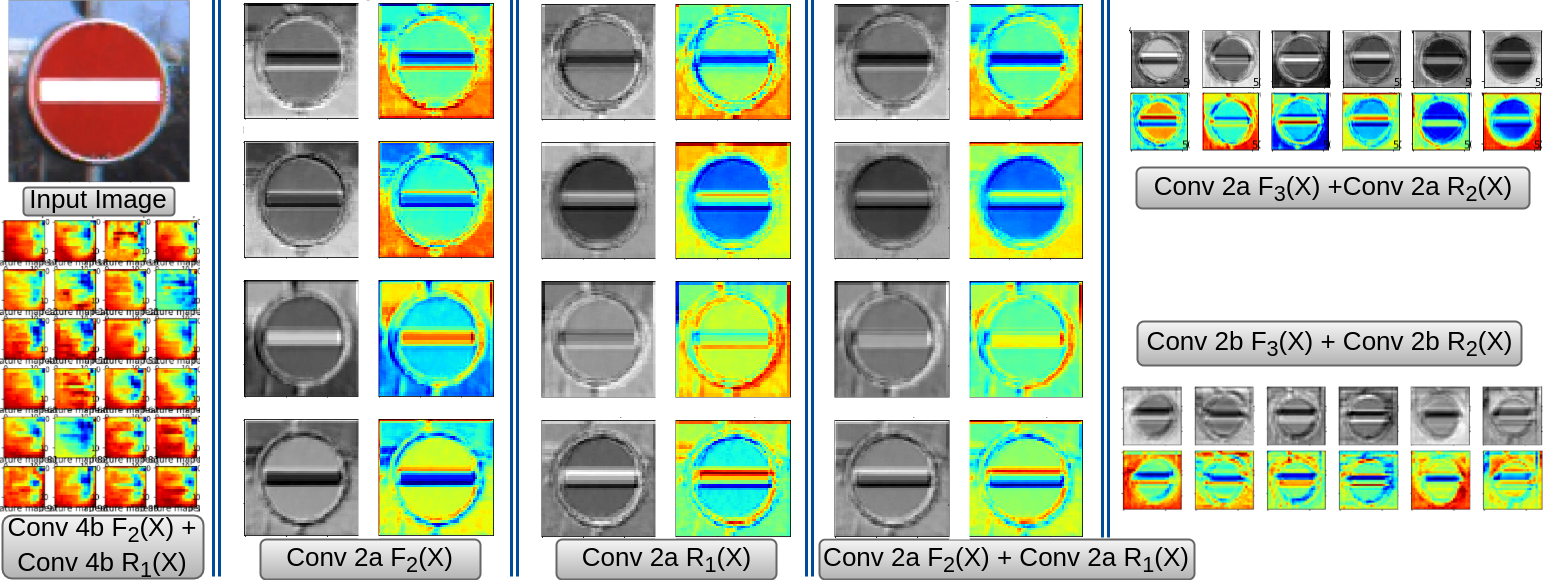}
\caption{View Point Variation and Feature Maps at Different Layers of our Dilated Inner Residual Deep Conv-Net. Each Layer's findings are shown using both activation scale and hit-maps. Terms like  Conv 2a \( F_{2}(X) \), Conv 2a \( R_{1}(X) \) used are inferred with juxtaposition of notations used in Figure ~\ref{fig2} and Table \ref{tab2}.}
\label{fig5}
\end{figure*}

\subsection{Observation and Justification}
As Figure ~\ref{fig5} demonstrates, the \( 2^{nd} \) consecutive [\( k_{3} \)] set of kernels in the \( 1^{st} \) residual block (Conv 2a \( F_{2}(X) \)) succeed at finding round and straight edges whereas, the \( 1^{st} \) dilated [\( k_{3}, r=2 \)] set of kernels in the very same block (Conv 2a \( R_{1}(X) \)) detects noise around those edges thus leaving the findings blunt. However when added together (Conv 2a \( F_{2}(X) \) + Conv 2a \( R_{1}(X) \)), they tend to trigger more robust features like precise edges and shades. Concurrently, deeper layers (Conv 2a \( F_{3}(X) \) + Conv 2a \( R_{2}(X) \)) succeed at rendering the background details obsolete thus focusing further on the traffic signs rather than the background. Subsequently, the feature maps toward the end layers (Conv 4b \( F_{2}(X) \) + Conv 4b \( R_{1}(X) \)) extract very abstract level details with each iteration of optimization. On the other hand, we also trained various large scale Image-net challenge winning image classifiers to compare our results, the details of which is shown in Table \ref{tab1}.

\begin{figure}[bp]
\centering
\includegraphics[width=8.5cm,height=3.1cm]{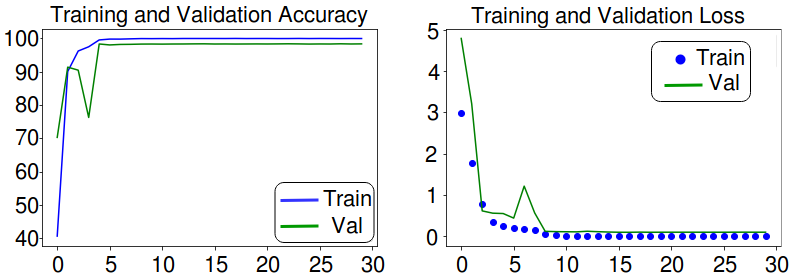}
\caption{Training and Validation Accuracy and Loss: GTSRB}
\label{fig6}
\end{figure}

\section{Conclusion}
As many traffic sign recognition CNN models as there exist, ours evidently requires the least amount of learn-able parameters and training time. Furthermore, our CNN model is the first of it's kind to be robust enough across different traffic sign recognition or classification benchmarks. With all the tech giants focusing more on producing self driving cars, such one-for-all as well as lightning fast traffic sign recognition platform would come in handy with numerous business values in perspective. Needless to say, when it comes to driving, understanding of the environment at precision level is the most important aspect. Consequently, faster recognition of the objects in the environment- of which traffic signs are an inevitable part of, determines how accurately a self driving car agent or a human driver for that matter would actuate in the environment. Furthermore, the devised methodologies and theories behind the making of this CNN architecture opens up possibilities of re-visioning, rethinking and re-scaling in large-scale image classification which sure enough, is a domain for further research in the distant future.

\section*{Acknowledgment}
We would like to thank \href{http://ccse.iub.edu.bd/}{``Center  for  Cognitive  Skill  Enhancement" Lab} at Independent University Bangladesh for both the technical and inspirational supports provided.

\bibliographystyle{IEEEtran}
\bibliography{reference}

\end{document}